\documentclass[conference]{IEEEtran}
\IEEEoverridecommandlockouts
\usepackage{cite}
\usepackage{amsmath,amssymb,amsfonts}
\usepackage{algorithmic}
\usepackage{graphicx}
\usepackage{textcomp}
\usepackage{xcolor}

\usepackage{multirow}
\usepackage{hyperref}
\usepackage{colortbl}  
\usepackage{array}   
\usepackage[ruled]{algorithm2e}
\newcommand{\figref}[1]{Figure~\ref{#1}}
\newcommand{\tabref}[1]{Table~\ref{#1}}
\newcommand{\equref}[1]{Eq.~\ref{#1}}
\newcommand{\aloref}[1]{Algorithm.~\ref{#1}}

\def\BibTeX{{\rm B\kern-.05em{\sc i\kern-.025em b}\kern-.08em
    T\kern-.1667em\lower.7ex\hbox{E}\kern-.125emX}}
\begin{document}

\title{The Devil is in the Conflict: Disentangled Information Graph Neural Networks For Fraud Detection}

\author{\IEEEauthorblockN{Zhixun Li\textsuperscript{1,$\dag$}, Dingshuo Chen\textsuperscript{2,3,$\dag$}, Qiang Liu\textsuperscript{2,3}, Shu Wu\textsuperscript{2,3,*} \thanks{\textsuperscript{$\dag$}The first two authors contributed equally to this work.} \thanks{\textsuperscript{*}Corresponding author.}}
\IEEEauthorblockA{\textsuperscript{1}\textit{School of Computer Science and Technology, Beijing Institute of Technology} \\
\textsuperscript{2}\textit{Center for Research on Intelligent Perception and Computing, National Laboratory of Pattern Recognition,} \\
\textit{Institute of Automation, Chinese Academy of Sciences} \\
\textsuperscript{3}\textit{School of Artificial Intelligence, University of Chinese Academy of Sciences}\\
lizhixun@bit.edu.cn, dingshuo.chen@cripac.ia.ac.cn, \{qiang.liu, shu.wu\}@nlpr.ia.ac.cn}
}

\maketitle

\begin{abstract}
Graph-based fraud detection has heretofore received considerable attention. Owning to the great success of Graph Neural Networks (GNNs), many approaches adopting GNNs for fraud detection has been gaining momentum. However, most existing methods are based on the strong inductive bias of homophily, which indicates that the context neighbors tend to have same labels or similar features. In real scenarios, fraudsters often engage in camouflage behaviors in order to avoid detection system. Therefore, the homophilic assumption no longer holds, which is known as the inconsistency problem. In this paper, we argue that the performance degradation is mainly attributed to the inconsistency between topology and attribute. To address this problem, we propose to disentangle the fraud network into two views, each corresponding to topology and attribute respectively. Then we propose a simple and effective method that uses the attention mechanism to adaptively fuse two views which captures data-specific preference. In addition, we further improve it by introducing mutual information constraints for topology and attribute. To this end, we propose a \textbf{D}isentangled \textbf{I}nformation \textbf{G}raph \textbf{N}eural \textbf{N}etwork (DIGNN) model, which utilizes variational bounds to find an approximate solution to our proposed optimization objective function. Extensive experiments demonstrate that our model can significantly outperform state-of-the-art baselines on real-world fraud detection datasets.
\end{abstract}


\section{Introduction}
Graph-based fraud detection is a crucial task and has tremendous impact in various applications, such as opinion fraud detection \cite{li2019spam}, fake news detection \cite{dou2021user, xu2022mining}, review spams\cite{deng2022markov} and financial fraud detection \cite{wang2019semi, liu2021pick}. In these scenarios, as graph can effectively model the correlations among entities, interactive activities on platform can be characterized as a graph, where users or objects are often treated as nodes, and transactions or relations between them are treated as edges.

\begin{figure}[t]
\centering
\includegraphics[width=\linewidth]{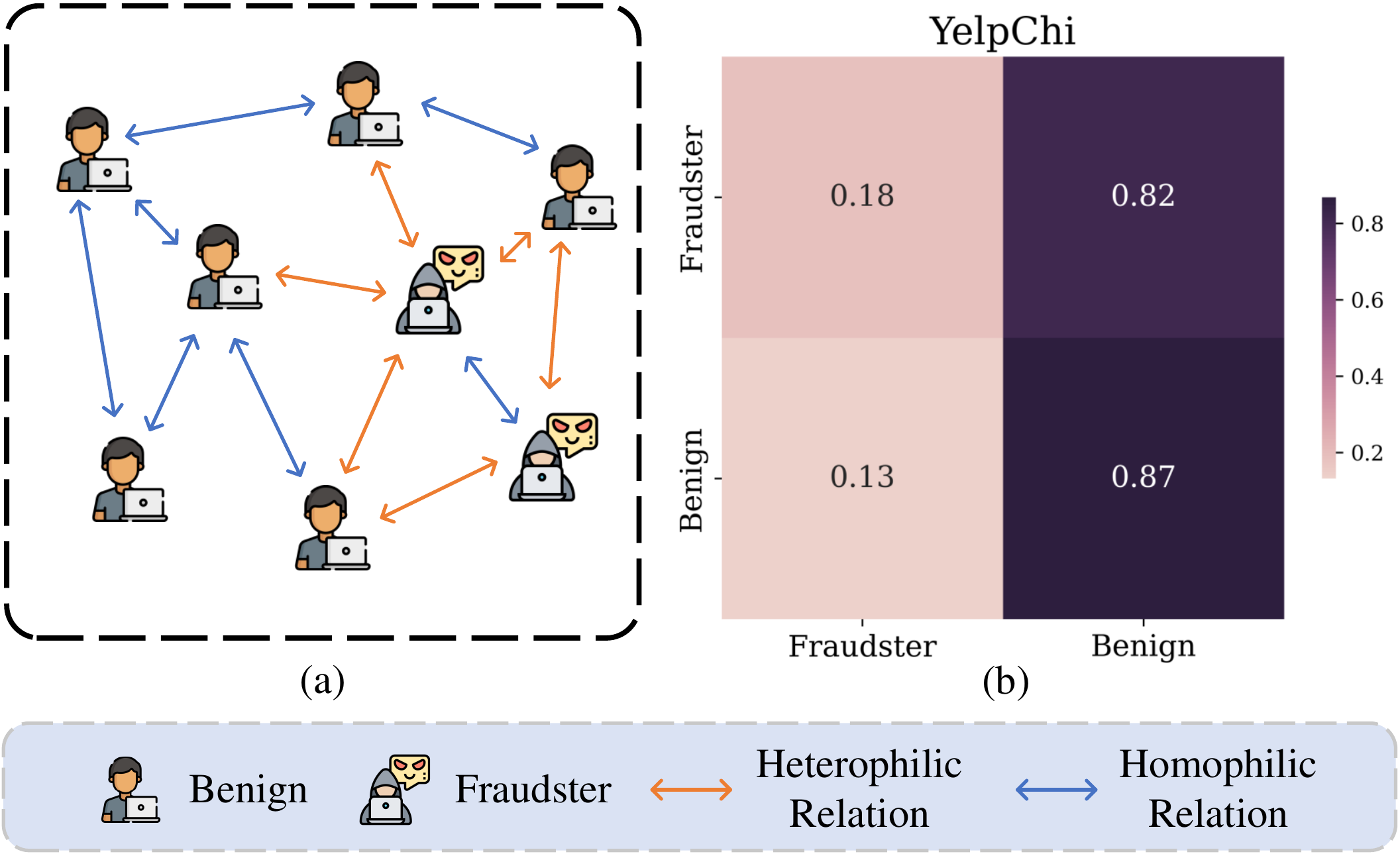}
\caption{(a) Illustration of graph-based fraud detection. (b) Neighbor distribution of fraudsters and benign users in YelpChi dataset.}
\label{fig:example}
\end{figure}

Numerous techniques have been proposed to detect the fraudsters. Recently, driven by the powerful representation capability of graph structure and advances of Graph Neural Networks (GNNs) \cite{kipf2016semi, hamilton2017inductive, velickovic2017graph}, many approaches try to harness GNNs for fraud detection on either homogeneous or heterogeneous graphs. The main idea is to leverage GNNs to learn expressive node representations with the goal of distinguishing abnormal nodes from the normal ones in the latent embedding space. Message-Passing GNNs (MP-GNNs) are mainstreaming in recent years, which aggregate neighbor node features and achieve local smoothing by stacking layers. Although MP-GNNs can obtain satisfactory performance on most of cases, the strong inductive bias of homophily limits their representative ability on heterophilic graphs. Some works \cite{nt2019revisiting} point out that plentiful GNNs can be seen as low-pass filters, so their generalization ability on high frequency graph signals are poor. In fraud detection task, fraudsters often imitate normal users in order to camouflage themselves, hence they will interact with normal users more frequently. For instance, normal users account for 81\% of the fraudsters' neighbor nodes in YelpChi dataset (\figref{fig:example}). In other words, fraudsters' features are inconsistent with their behaviors (interactions, e.g., topological structure). Thus, recall that MP-GNNs do not work well on heterophilic graphs, they fail to tackle the inconsistency phenomenon in graph-based fraud detection and fraudsters could fool the detection system.


Recently, a few works have noticed this problem, and they employ aggregating weights to reduce the adverse impact of dissimilar neighbors, or set similarity-aware thresholds to select and re-link similar nodes. For instance, GraphConsis \cite{liu2020alleviating} computes consistent score between connected node pairs as the sampling probability. PC-GNN \cite{liu2021pick} combines label information and latent embeddings as distance function to measure similarity. Although such methods can alleviate the inconsistency problem in some extent, they discard a lot of information during filting dissimilar neighbors out, thus they may lead to sub-optimal performance.


In this paper, we analyze the inconsistency problem in graph-based fraud detection task, which has been obstructing a full understanding of this field. First, we clarify that the inconsistency problem is the bottleneck of graph fraud detection.  According to \cite{ma2021unified}, the underlying optimization process of GNNs is equivalent with minimizing the topology and attribute constraints, and Yang et al.\cite{yang2022graph} indicates that the degradation of performance is imputed to the compromise between topology and attribute. Due to the camouflage behaviors (topology) of fraudsters, which are inconsistent with their essence (attribute), this conflict in fraud networks may injure the discriminative ability of GNNs. Second, the forefronts of different datasets are diverse, and most existing methods are not satisfactory in fusing topological structures and node attributes \cite{wang2020gcn}. For example, fraudsters may possess distinguishable attribute on some platforms, but their deceptive behaviors can confuse the detection model. Therefore, we are motivated to explore a novel method that is able to minimize the conflict between topology and attribute and meanwhile effectively extract most task-relevant information from datasets. 

We borrow the concept of multi-view learning problems to graph-based fraud detection task and propose a simple and effective model, \textbf{D}isentangled \textbf{I}nformation  \textbf{G}raph \textbf{N}eural \textbf{N}etworks (DIGNN). Technically, we first disentangle fraud networks into topology and attribute views. Next, we employ attention mechanism to fuse two view embeddings adaptively for extracting task-relevant information. Surprisingly, we observe that this simple method surpasses all state-of-the-art baselines. This empirically proves that the conflict between topology and attribute causes the inconsistency problem. Besides, to further decrease the entanglement between topology and attribute and improve the performance, we design a new optimization objective based on information theory, which resorts to variational bounds to minimize mutual information between two views and maximize the mutual information between view embeddings and original inputs.

We conduct extensive experiments to compare our proposed model with existing graph-based fraud detection models, the results demonstrate the effectiveness of our model. In summary, the contributions of this paper can be summarized as follows:
\begin{itemize}
\item We analyze the cause of the inconsistency problem, and point out that it is mainly attributed to the conflict between topology and attribute. In light of this, we propose a simple yet effective model, DIGNN, which firstly disentangles fraud network into two views and fuses them by attention mechanism.
\item We propose a novel optimization objective based on mutual information theory and theoretically derive its upper bound for tractable calculation.
\item We verify the effectiveness of our model on real-world fraud detection datasets. It is shown that our model is able to significantly improve the performance in terms of all commonly adopted metrics.
\end{itemize}

\section{Related Work}
\subsection{Graph-based Fraud Detection}
The core idea of graph-based fraud detection task is taking the advantages of GNNs to get the discriminative node embeddings, and find out the malicious ones in the latent space. Examples include \cite{liu2020alleviating, dou2020enhancing, wang2021decoupling} for review fraud detection, \cite{dou2021user, xu2022mining} for fake news detection and \cite{wang2019semi, liu2021pick, ao2021temporal, liang2021credit, zhang2022efraudcom} for financial fraud detection. Ma et al. \cite{ma2021comprehensive} provides a comprehensive investigation on graph-based fraud detection.

Most of existing GNNs methods holds homophilic assumption that neighbor nodes share same labels or similar features. However, fraudsters will try to conceal themselves, so that their features are inconsistent with their camouflage behaviors. Some graph-based fraud detection works have noticed this problem. GraphConsis \cite{liu2020alleviating} pioneers to formulate and tackle the inconsistency problem. They introduce three kinds of inconsistency phenomenon existing in fraud networks. CARE-GNN \cite{dou2020enhancing} devises a label-aware similarity measure to find informative neighboring nodes and utilizes reinforcement learning to select similar neighbors. FRAUDRE \cite{zhang2021fraudre} aggregates difference between adjacent node pairs. PC-GNN \cite{liu2021pick} devises a choose operation to select beneficial neighbors based on feature similarity. IHGAT \cite{liu2021intention} is devised to encode both sequence-like intentions and relationship among transactions for leveraging the cross-interaction information.

Our model is different from all above works. We innovatively disentangle topology and attribute and consider graph learning as a multi-view learning problem, instead of measuring similarity between adjacent node pairs.
\subsection{Multi-view on GNNs}
Topology and attribute are two essential compositions of graphs. However existing state-of-the-art GNN models are disable to effectively fuse topological structure and node attributes. AM-GCN \cite{wang2020gcn} uses $k$-nearest neighbor to construct feature graph and combine it with topological structure view and common embeddings. SCRL \cite{liu2021self} designs a self-supervised approach to maximize the agreement of the embeddings in the topology graph and the feature graph. A recent work \cite{yang2022graph} claims that the interference between topology and attribute is mainly ascribed to compromises between them. LINKX \cite{lim2021large} processes node attributes and topological structure in an orthogonal manner. In this paper, we also follow this idea and extend it by proposing a novel architecture and optimization objective.

Information-theoretic methods have been gaining momentum in recent years, which take into consideration the mutual dependency of different views. MIB\cite{Federici2020} extends the information bottleneck principle to unsupervised multi-view setting to discard superfluous information. DVIB\cite{bao2021} and CMIB\cite{wan2021} leverage mutual information constrains to better preserve shared and private information of multi-view learning. To cope with intractable computation of mutual information, these methods adopt variational inference to optimize objective lower and upper bounds. In comparison, our model propose a novel optimization objective to  reconcile consistency and complementarity between topology and attribute views. Equipped with variational inference, we also approximate the mutual information with derived bounds.

\section{PRELIMINARIES}


\subsection{Problem Statement}
\textbf{Definition 1. Graph-based Fraud Detection.} Given a fraud network $ \mathcal{G}=(\mathcal{V},\mathbf{A},\mathbf{X}) $, where $ \mathcal{V}=\{v_1,v_2,\ldots,v_N\} $ is the set of nodes; $ \mathbf{A}\in\mathbb{R}^{N\times N} $ is the adjacency matrix, if $ v_i $ and $ v_j $ are connected, $ \mathbf{A}_{ij}=1 $, otherwise, $ \mathbf{A}_{ij}=0$; $ \mathbf{X}\in\mathbb{R}^{N\times D} $ denotes node feature matrix, each node $ v_i $ is associated with a $D$-dimensional feature vector $ \mathbf{x}_i $ and a label $ y_i\in\{0,1\} $, where 0 denotes the node is a normal user (negative) and 1 indicates it is a fraudster (positive). The core idea of graph-based fraud detection is to learn discriminative node embeddings to detect the anomaly samples in latent space.

\textbf{Definition 2. Graph Neural Networks.} Most of GNNs follow message passing mechanism which uses feature transformation and aggregation operations to capture the structural and attribute information. One of the most popular and representative GNNs model is graph convolutional networks (GCNs). The forward inference at the $ l $-th layer of GCNs is formally defined as:
\begin{equation}\label{gcn}
\mathbf{H}^{(l)}=\sigma(\hat{\mathbf{A}}\mathbf{H}^{(l-1)}\mathbf{W}^{(l)}),
\end{equation}
where $ \sigma(\cdot) $ is nonlinear activation function, $ \mathbf{W}^{(l)}\in\mathbb{R}^{d\times d} $ is the linear transformation matrix, $ \mathbf{H}^{(l)} $ denotes the node embedding matrix at the $ l $-th layer, and $ \mathbf{H}^{(0)} = \mathbf{X}$, $\hat{\mathbf{A}}$ is the normalized adjacency matrix, which can be implemented by row-normalization, $\hat{\mathbf{A}}=\mathbf{D}^{-1}(\mathbf{A}+\mathbf{I})$ or symmetric normalization, $\hat{\mathbf{A}}=\mathbf{D}^{-\frac{1}{2}}(\mathbf{A}+\mathbf{I})\mathbf{D}^{-\frac{1}{2}}$, and $ \mathbf{D} $ is a diagonal matrix, $ \mathbf{I} $ is an identity matrix.

Interestingly, some works \cite{ma2021unified} have declared that representative GNN models can be regarded as solving a \textbf{Graph Signal Denoising} problem, which given a noisy signal $\mathbf{S}\in\mathbb{R}^{N\times d_{in}}$ on graph $\mathcal{G}$, the goal is to recover a clean signal $\mathbf{F}\in\mathbb{R}^{N\times d_{out}}$:
\begin{equation}\label{gcn}
\underset{\mathbf{F}}{\arg\min}\ \mathcal{L}=\lVert\mathbf{F}-\mathbf{S}\rVert_F^2+c\cdot tr(\mathbf{F}^\top\mathbf{L}\mathbf{F}),
\end{equation}
where the first term guides output signal $\mathbf{F}$ similar to original signal $\mathbf{S}$ and the second term encourages signal smoothness on graph.


\begin{figure*}[t]
\centering
\includegraphics[width=\textwidth]{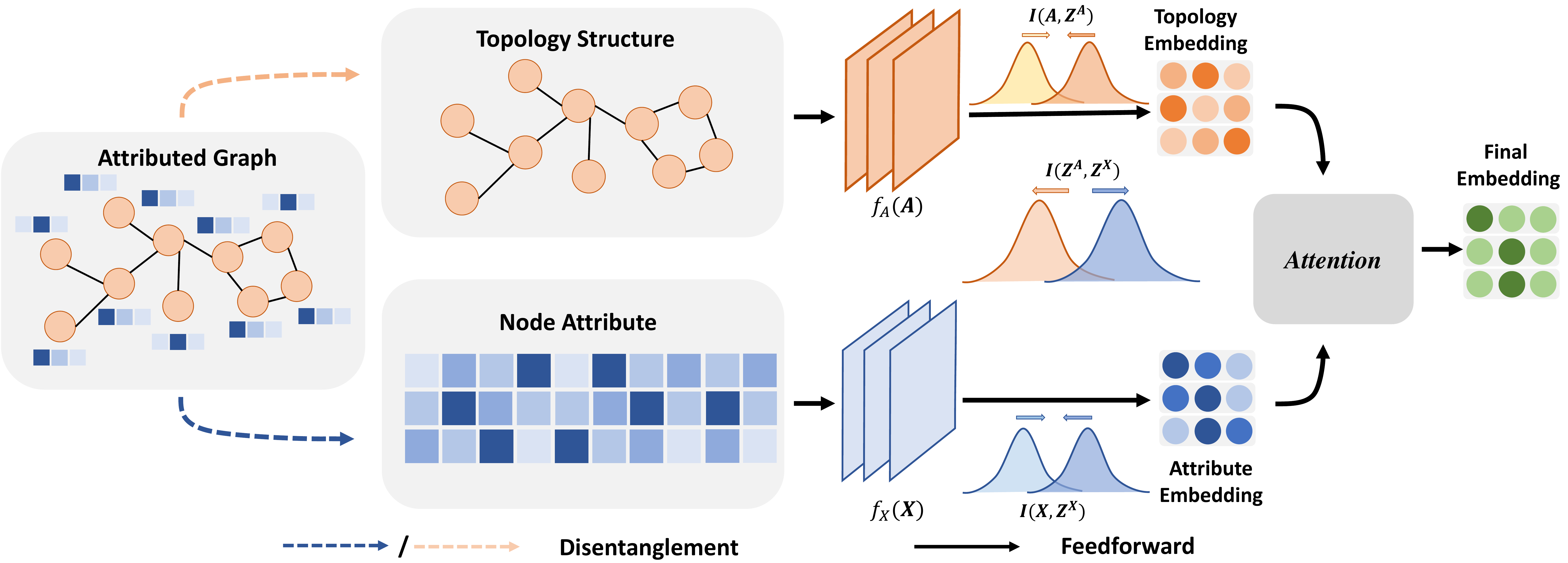}
\caption{The overview of our proposed DIGNN. The attributed fraud network is disentangled into topological structure and node attributes. DIGNN processes these two views in parallel and fuses them by attention mechanism. In addition, in order to further reduce the entanglement between two views, DIGNN minimizes the mutual information between topology embeddings and attribute embeddings, and maximizes the mutual information between embeddings and input data respectively.}
\label{fig:overview}
\end{figure*}

\section{METHOD}
In this section, we will present our model, Disentangled Information Graph Neural Network (DIGNN). \figref{fig:overview} gives an overview of our model. It consists of three main objectives: 1. Disentangle attribute fraud network into topology and attribute views and fuse them by attention mechanism, the final embeddings are trained with Cross-Entropy loss; 2. To further reduce the conflict between two views, we minimize the mutual information between them; 3. In order to maintain the semantic information from input space, we maximize the mutual information between view-specific embeddings and their original inputs.

\subsection{View-specific Embedding}
It is universally acknowledged that topology and attribute are of vital importance for graph learning. However, in graph fraud detection scenario, traditional message passing along neighboring nodes is inappropriate as graph signal smoothing makes fraudsters more indistinguishable. To alleviate the inconsistency problem, we disentangle the topology and attribute information and encode them in parallel.

Given an attributed fraud network $ \mathcal{G} $, it can be disentangled into topology view $ \mathbf{A} $ and attribute view $ \mathbf{X} $. Here we provide two view encoders $ f_A, f_X $ for each input view, as shown in \figref{fig:overview}. Specifically, we employ Multi-Layer Perceptron (MLP) as encoders to obtain view-specific embeddings $ \mathbf{Z}^A, \mathbf{Z}^X\in\mathbb{R}^{N\times d} $:
\begin{equation}\label{encoder}
\mathbf{Z}^A = f_A(\mathbf{A}),\ \mathbf{Z}^X = f_X(\mathbf{X}),
\end{equation}
in which $ d $ is the embedding dimension. With these two embeddings, we need to fuse them to obtain final representation and extract task-relevant information.

\subsection{Cross-view Fusion}

Now we have two view-specific embeddings $\mathbf{Z}^A$ and $\mathbf{Z}^X$, we then perform cross-view fusion by utilizing attention mechanism. The attention value $\omega_i$ can be represented as:
\begin{equation}
    \omega_i^j=q\cdot\tanh(\mathbf{W}\cdot(\mathbf{z}_i^j)^\top+b),\ j\in\{A,X\}
\end{equation}
where $q$ denotes the learnable attention vector, $\mathbf{W}$ is the weight matrix and $b$ is bias vector. Thus, we can get the attention values $\omega^A_i$ and $\omega^X_i$ for view-specific embeddings $\mathbf{z}^A_i$ and $\mathbf{z}^X_i$, respectively. Then we normalize them via softmax function to get the final weight:
\begin{equation}
    \alpha_i^j=\text{softmax}(\omega_i^j)=\frac{\exp(\omega_i^j)}{\exp(\omega_i^A)+\exp(\omega_i^X)},\ j\in\{A,X\}
\end{equation}
Larger attention weight $\alpha_i$ implies that the corresponding embeddings is more important, and it is determined by specific dataset. Then the final output embedding $\mathbf{z}_i$ can be combined by two view-specific embeddings with its corresponding attention weight as:
\begin{equation}\label{attention}
    \mathbf{z}_i=\alpha_i^A\cdot\mathbf{z}_i^A+\alpha_i^X\cdot\mathbf{z}_i^X.
\end{equation}
And we put it into a linear classifier, while training by a cross-entropy loss function:
\begin{equation}\label{crossentropy}
    \mathcal{L}_{ce}=\sum_{v_i\in\mathcal{V}_\text{train}}-\log(y_i\cdot\sigma(\mathbf{W}'\cdot \mathbf{z}_i+b'))
\end{equation}
in which $\mathbf{W}'$ and $b'$ is the weight matrix and bias vector of linear classifier, $\sigma$ is a softmax function, and $\mathcal{V}_\text{train}$ is the training node set.

\subsection{Mutual Information Optimization}
Up to now, we have discussed how to get view-specific embeddings and fuse them with attention mechanism. However, as mentioned in \cite{yang2022graph}, the representative GNN models tend to deteriorate their expressive power due to interference between attribute and topology. In spite of decoupling operation, it is still impractical to look forward to injecting mutual-exclusive learning ability to our model simply equipped with attention mechanism. In other words, we need to seek some principles to guide the training procedure. By leveraging the information theory, we propose a novel optimization objective to alleviate the aforementioned problem. Furthermore, we derive the variational bound of our optimization objective and discuss the intrinsic effect and intuitive insight.
Without loss of generality, we let $\mathbf{X}_1, \mathbf{X}_2$ to represent original views and $\mathbf{Z}_1, \mathbf{Z}_2$ to represent view-specific embeddings for ease of reading. 

\subsubsection{\textbf{Optimization principles}}
The first principle aims to induce model to learn mutual-exclusive embeddings, which ameliorates the compromise problem between attribute and topology. Considering that mutual information measures the mutual dependence of variables, we introduce the constraint term $\textbf{min } I(\mathbf{Z}_1,\mathbf{Z}_2)$ to our optimization objective. In this way, model is able to reduce the redundancy and enhance the ability on exploiting sufficient semantic information in embedding space with limited dimensionality.

Nevertheless, mutual-exclusive constraint is prone to impair the helpful shared information. For instance, in Amazon dataset, handcrafted features are highly correlated to social networks (topology), thus mutual-exclusive constraint will injure attribute semantics during training. The  second principle builds the relationship between view-specific embeddings and their original inputs. In virtue of rich but distinct semantics inherent in the attribute and topology, it is necessary to extract useful features and meanwhile maintain respective information from input data space. We further introduce the constraint term $\textbf{max } I(\mathbf{Z}_i, \mathbf{X}_i)$ to our optimization objective to encode inputs with more view-specific information available. To sum up, our mutual information optimization objective can be summarized as follow:
\begin{align} 
\text{min }I(\mathbf{Z}_1, \mathbf{Z}_2)-\sum_{i=1}^2 I(\mathbf{Z}_i,\mathbf{X}_i)
\end{align}

\subsubsection{\textbf{Theoretical Analysis}}
Recall that the optimization objective has the form
$\text{min }I(\mathbf{Z}_1, \mathbf{Z}_2)-\sum_{i=1}^2 I(\mathbf{Z}_i,\mathbf{X}_i)$. However, it is intractable to directly calculate the mutual information for high dimensional variables \cite{Belghazi2018}. We alternatively sort to derive variational bounds of the mutual information to find an approximate solution to original objective function.
In this work, $\mathbf{X},\mathbf{Y},\mathbf{Z}$ denote random variables, and $x,y,z$ denote corresponding instances. 

\textbf{Lower bound of $I(\mathbf{Z}, \mathbf{X})$}.
For conciseness, we use term $I(\mathbf{Z}, \mathbf{X})$ to represent $I(Z_i,X_i)(i\in \{1, 2\})$.  According to the definition of mutual information, we have
\begin{align}
I(Z, X)&=\mathbb{E}_{p(x,z)}\log\frac{p(x|z)}{p(x)}\\
       &\ge \mathbb{E}_{p(x,z)}\log\ q(x|z)+H(\mathbf{X}) \label{Izx}
\end{align}
where $H(\mathbf{X})$ is the entropy of $X$; $q(x|z)$ is the variational approximation of conditional distribution $p(x|z)$. Notice that the entropy of random variable $X$ is independent of our optimization procedure. Therefore, maximization of $I(\mathbf{Z}, \mathbf{X})$ is equivalent to maximize  $\mathbb{E}_{p(x,z)}log\ q(x|z)$. 

\begin{figure}[t]
\centering
\includegraphics[width=\linewidth]{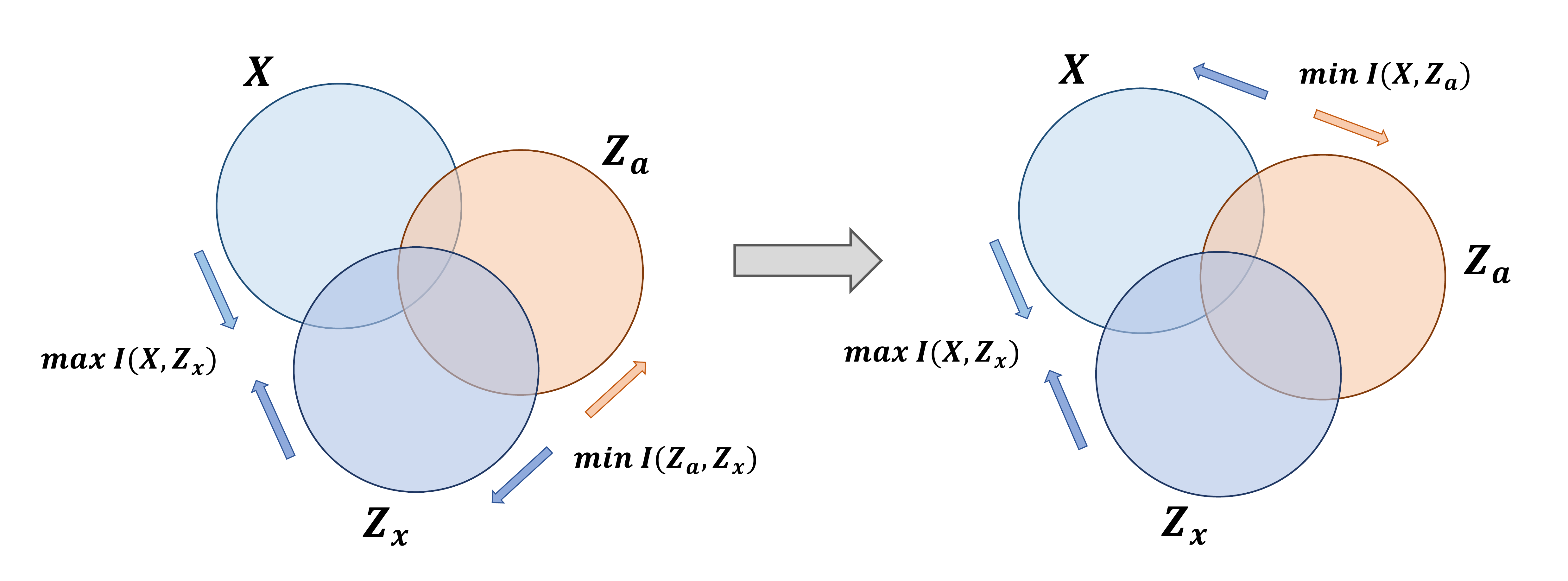}
\caption{A demonstration of original optimization objective (\emph{left}) and lower bound target (\emph{right}).}
\label{fig:mutual}
\end{figure}

\textbf{Upper bound of $I(\mathbf{Z}_1, \mathbf{Z}_2)$}.
The overall optimization transformation is demonstrated as \figref{fig:mutual}. At the beginning, we introduce some basic properties in information theory \cite{Cover1991}:
\begin{gather}
I(\cdot ; \cdot) \ge 0 \\
I(\mathbf{X}\mathbf{Y};\mathbf{Z}) = I(\mathbf{Y};\mathbf{Z})+I(\mathbf{X};\mathbf{Z}|\mathbf{Y})
\end{gather}

Following \cite{Federici2020}, we have
\begin{align*} \label{eqs3}
I(\mathbf{Z}_1,\mathbf{Z}_2)  & \le I(\mathbf{Z}_1,\mathbf{Z}_2) + I(\mathbf{Z}_1;\mathbf{X}_2|\mathbf{Z}_2) \\
            &= I(\mathbf{Z}_1;\mathbf{Z}_2\mathbf{X}_2) \\
            &= I(\mathbf{Z}_1;\mathbf{Z}_2\mathbf{X}_2)-I(\mathbf{Z}_1;\mathbf{Z}_2|\mathbf{X}_2)\\
            &= I(\mathbf{Z}_1, \mathbf{X}_2)
\end{align*}
Here we assume that $\mathbf{Z}_2$ is a sufficient representation of $\mathbf{X}_2$, which means $I(\mathbf{Z}_1;\mathbf{Z}_2|\mathbf{X}_2)=0$ holds. According to symmetry of mutual information, we can follow the same formulation and derive an equivalent form of $I(\mathbf{Z}_1,\mathbf{Z}_2)$.

\begin{align} 
I(\mathbf{Z}_2,\mathbf{Z}_1) \le I(\mathbf{Z}_2, \mathbf{X}_1)
\end{align}

Then, we take the mutual information $I(\mathbf{Z}_1, \mathbf{X}_2)$ as an example and derive the upper bound. We have:
\begin{align}
I(\mathbf{Z}_1, \mathbf{X}_2)&=\mathbb{E}_{p(z_1,x_2)}\log\frac{p_{x_1}(z_1|x_2)}{p(z_1)}\\
      &=\mathbb{E}_{p(z_1,x_2)}log\frac{p_{x_1}(z_1|x_2)}{r(z_1)}  \frac{r(z_1)}{p(z_1)}\\
      &=\mathbb{E}_{p(z_1,x_2)}log\frac{p_{x_1}(z_1|x_2)}{r(z_1)} - 
      KL \left[r(z_1) \| p(z_1) \right]\\
       &\le \mathbb{E}_{p(z_1,x_2)}\log\frac{p_{x_1}(z_1|x_2)}{r(z_1)}\\
       &=\mathbb{E}_{p(z_1,x_2)}\log\frac{p_{x_1}(z_1|x_2)}{p_{x_2}(z_2|x_2)}\\
       &\qquad\qquad\qquad+\mathbb{E}_{p(z_1,x_2)}\log\frac{p_{x_2}(z_2|x_2)}{r(z_1)}
\end{align}
where $p_{x_1}$ and $p_{x_2}$ represent encoders that encode information from original feature space. The upper bound will become tighter as the marginal distribution $r(z_1)$ approaches the priors $p(z_1)$. By observing the two terms in the upper bound. The first term measures the difference of two latent representations from $p_{x_1}$ and $p_{x_2}$ but with the same input $x_2$, while the second term measures the difference between encoder $p_{x_2}$ with the approximated margin $r(z_1)$. According to \cite{bao2021}, the two terms has the same optimization directions, thus we simplify the upper bound to 
\begin{align} 
I(\mathbf{Z}_1,\mathbf{X}_2) \le \mathbb{E}_{p(z_1,x_2)}\log\frac{p_{x_2}(z_2|x_2)}{r(z_1)}
\end{align}

Again, we take into consideration the symmetry of the mutual information and take average of the two form to formulate the final upper bound. 
\begin{equation} 
\begin{split}
I(\mathbf{Z}_1,\mathbf{Z}_2) \le \frac{1}{2} 
&\left[\mathbb{E}_{p(z_1,x_2)}\log\frac{p_{x_2}(z_2|x_2)}{r(z_1)} \right.\\
& +\left.\mathbb{E}_{p(z_2,x_1)}\log{\frac{p_{x_1}(z_1|x_1)}{r(z_2)}}\right]
\end{split}
\end{equation}

In practice, we minimize the reconstruction loss to equivalently minimize the lower bound of $I(\mathbf{Z},\mathbf{X})$, as done in auto-encoder models\cite{vincent2010stacked}. According to the type of input $x$, $q(x|z)$ can be any appropriate distribution with known probability density function. Here we let the $q(x|z)$ be the Gaussian distribution $\mathcal{N}(x;\mu(z),\sigma^2I)$ with given variance $\sigma^2$ and deterministic mean function $\mu(\cdot)$ which is parameterized by neural networks, we have
\begin{align} 
\mathcal{L}_{rec}&= -\sum_{i=1}^2\mathbb{E}_{p(x_i,z_i)}\log\ q(x_i|z_i)\label{rec_loss} \\
 &\propto \sum_{i=1}^2\mathbb{E}_{p(x_i,z_i)}\left[\left\| x_i-\mu_i(z_i) \right\|_2^2\right]
\end{align}

To maximize the upper bound of $I(\mathbf{Z}_1,\mathbf{Z}_2)$, we define $p_{x_i}(z_i|x_i)$ and $r(z_i)$ as the Gaussian distribution $\mathcal{N}(z_i;\mu(x_i),\sigma_i^2I)$ and $\mathcal{N}(z_i;\mu_i,\sigma_i^2I)$ respectively, where $\mu_i$ and $\sigma_i$ are given expectation and variance. Then we employ the reparameterization trick\cite{kingma2013auto} to rewrite $p_{x_i}(z_i|x_i)=p(\epsilon_i)$, where $z_i=\mu(x_i)+\epsilon_i\sigma,\epsilon_i \sim \mathcal{N}(0,I)$ and our mutual-exclusive loss $\mathcal{L}_{exc}$ is defined as
\begin{align} 
\mathcal{L}_{exc} 
&=\frac{1}{2} \left[\mathbb{E}_{p(z_1,x_2)}\log\frac{p_{x_2}(z_2|x_2)}{r(z_1)}+
\mathbb{E}_{p(z_2,x_1)}\log\frac{p_{x_1}(z_1|x_1)}{r(z_2)}\right] \label{exc_loss} 
\end{align}

Eventually, the overall optimization objective is formulated as follow
\begin{align} 
\mathcal{L} = \mathcal{L}_{ce} + \alpha\cdot\mathcal{L}_{rec} + \beta\cdot\mathcal{L}_{exc} \label{final_loss}
\end{align}
where $\alpha$ and $\beta$ are scalar factors. Moreover, it is worth noting that the second term reconstruction loss is equivalent to graph signal denoising but without signal smoothness, which is reasonable considering the inconsistency problem of graph anomaly detection. Intuitively, our loss function denoises the original graph signal and achieves mutual exclusion between attribute and topology together with supervised information. The training procedure is presented in \aloref{dignn_algo}.

\begin{algorithm}[]  
    \label{dignn_algo}
	\caption{DIGNN: Disentangled Information Graph Neural Network}
	\LinesNumbered 
	\KwIn{$\mathcal{G}=(\mathcal{V},\mathbf{A},\mathbf{X})$: A fraud network, $\mathcal{V}_\text{train}$: Set of training nodes, $N_\text{epoch}$: Number of total training epochs, $N_\text{batch}$: Number of training batch size, $d$: Dimension of hidden embeddings, $\alpha$ and $\beta$: hyperparameters of loss balance factors.}
	\KwOut{The vector representations for each node in $\mathcal{V}$}
	Initialization $v\in\mathcal{V}_\text{train}$\; 
	\For{$e=1,\ldots,N_\text{epoch}$}{
	    Calculate the number of training batches $B=\lceil\frac{|\mathcal{V}_\text{train}|}{N_\text{batch}}\rceil$\;
	    Down-sample negative samples according to the number of positive samples\;
	    \For{$b=1,\ldots,B$}{
	        Gather nodes of batch $b$ along with edges between them to construct sub-graph $\mathcal{G}_b=(\mathcal{V}_b,\mathbf{A}_b,\mathbf{X}_b)$\;
	        $\mathbf{Z}^A, \mathbf{Z}^X\leftarrow$ \equref{encoder}\;
	        $\mathbf{Z}\leftarrow$ fuse two view-specific embeddings $\mathbf{Z}^A$ and $\mathbf{Z}^X$ w.r.t. \equref{attention}\;
	        $\mathcal{L}_{ce}\leftarrow$ \equref{crossentropy} \ \ \tcp{Cross-Entropy Loss}
	        $\mathcal{L}_{rec}\leftarrow$ \equref{rec_loss};  \ \ \tcp{Reconstruction Loss}
	        $\mathcal{L}_{exc}\leftarrow$ \equref{exc_loss}; \ \ \tcp{Mutual-exclusive Loss}
	        $\mathcal{L}\leftarrow$ computer final loss w.r.t. \equref{final_loss}\;
	        Back-propagation to update parameters;
	    }
	}
\end{algorithm}



\section{Experiments}

In this section, we investigate the effectiveness of DIGNN, and aim to answer the following research questions:
\begin{itemize}
    \item \textbf{RQ1}: Does disentangle operation benefit to inconsistency problem?
    \item \textbf{RQ2}: Does DIGNN outperform the state-of-the-art methods for graph-based fraud detection?
    \item \textbf{RQ3}: How do different components of DIGNN contributes to performance improvement in graph-based fraud detection task?
    \item \textbf{RQ4}: What is the performance of DIGNN with respect to different hyperparameters?
\end{itemize}

\subsection{Experiment Setup}
\subsubsection{Datasets.} Our proposed DIGNN model is evaluated on two real-world opinion fraud network datasets: YelpChi and Amazon. The YelpChi dataset \cite{rayana2015collective} collects hotel and restaurant reviews on Yelp.com online platform. The nodes of YelpChi dataset are reviews with 32 handcrafted features and the dataset includes three relations: 1) R-U-R that connects reviews posted by the same user, 2) R-S-R that connects reviews under the same product with the same star rating, 3) R-T-R that connects two reviews under the same product posted in the same month. The Amazon dataset \cite{mcauley2013amateurs} includes product reviews under the Musical Instrument category. The nodes in the graph of Amazon dataset are users with 25 handcrafted features and also contain three relations: 1) U-P-U that connects users reviewing at least one same product, 2) U-S-U that connects users having at least one same star rating within one week, 3) U-V-U that connects users with top 5\% mutual review text similarities (measured by TF-IDF) among all users. The statistic of datasets is shown in \tabref{tab:statistic}.

\subsubsection{Baselines.} We compare with several representative state-of-the-art models to verify the effectiveness of DIGNN in graph-based fraud detection.

\begin{itemize}
    \item \textbf{GCN} \cite{kipf2016semi}: graph convolutional network achieved by aggregating features in the neighborhood to generate node embeddings.
    \item \textbf{GAT} \cite{velickovic2017graph}: graph attention network aggregates the neighbors with different aggregation weights calculated by attention mechanism.
    \item \textbf{GraphSAGE} \cite{hamilton2017inductive}: an inductive GNN model samples the neighbor by connection information and aggregates features by stacking layers. 
    \item \textbf{DR-GCN} \cite{shi2020multi}: a dual-regularized graph convolutional network to handle multi-class imbalanced graph representation learning.
    \item \textbf{CARE-GNN} \cite{dou2020enhancing}: a fraud detection GNN model utilizes a similarity measure to enhance aggregation and reinforcement learning to obtain optimal selection count.
    \item \textbf{FRAUDRE} \cite{zhang2021fraudre}: a GNN model aggregates difference between neighbors and tackles with class imbalance.
    \item \textbf{PC-GNN} \cite{liu2021pick}: a state-of-the-art graph-based fraud detection method, which proposed pick and choose operations to alleviate inconsistency and class imbalance.
    \item \textbf{DIGNN$_{\setminus S}$}: a variant of DIGNN, which is trained on node feature and adjacency matrix directly without mini-batch sampling.
    \item \textbf{DIGNN$_{\setminus M}$}: a variant of DIGNN, which  removed the mutual information optimization from DIGNN.
\end{itemize}

\begin{table}[]
\caption{The Statistic of Datasets.}
\begin{tabular}{c|ccccc}
\hline
\multicolumn{1}{l|}{dataset} & \begin{tabular}[c]{@{}c@{}}Nodes\\ (fraud\%)\end{tabular}  & Relation                                                                     & \#Edges                                                                             & Class                                                                  & \#Class                                                     \\ \hline
YelpChi                      & \begin{tabular}[c]{@{}c@{}}45,954\\ (14.53\%)\end{tabular} & \textit{\begin{tabular}[c]{@{}c@{}}R-U-R\\ R-T-R\\ R-S-R\\ ALL\end{tabular}} & \begin{tabular}[c]{@{}c@{}}49,315\\ 573,616\\ 3,402,743\\ 3,846,979\end{tabular}    & \begin{tabular}[c]{@{}c@{}}Fraudster\\ Benign\end{tabular}             & \begin{tabular}[c]{@{}c@{}}6,677\\ 39,277\end{tabular}      \\ \hline
Amazon                       & \begin{tabular}[c]{@{}c@{}}11,944\\ (6.87\%)\end{tabular}  & \textit{\begin{tabular}[c]{@{}c@{}}U-P-U\\ U-S-U\\ U-V-U\\ ALL\end{tabular}} & \begin{tabular}[c]{@{}c@{}}175,608\\ 3,566,479\\ 1,036,737\\ 4,398,392\end{tabular} & \begin{tabular}[c]{@{}c@{}}Fraudster\\ Benign\\ Unlabeled\end{tabular} & \begin{tabular}[c]{@{}c@{}}821\\ 7,818\\ 3,305\end{tabular} \\ \hline
\end{tabular}
\label{tab:statistic}
\end{table}

\begin{table*}[htbp]
\caption{Performance Comparison on YelpChi and Amazon.}
\renewcommand\arraystretch{1.3}
\resizebox{\textwidth}{!}{
\begin{tabular}{cc|ccc|ccc}
\hline
\multirow{2}{*}{Method}   & Dataset & \multicolumn{3}{c|}{Yelpchi} & \multicolumn{3}{c}{Amazon} \\ \cline{2-8} 
                          & Metric  & F1-macro  & AUC     & GMean  & F1-macro & AUC    & GMean  \\ \hline
\multirow{7}{*}{Baselines} 
& GCN &   0.4929$\pm$0.0025   &    0.6274$\pm$0.0034     &    0.1886$\pm$0.0063    &    0.5461$\pm$0.0363      &    0.8328$\pm$0.0111    &     0.2570$\pm$0.0789   \\
& GAT &     0.4879$\pm$0.0230      &   0.5715$\pm$0.0029      &   0.1659$\pm$0.0789     &   0.6464$\pm$0.0387       &   0.8102$\pm$0.0179     &     0.6675$\pm$0.1345   \\
& GraphSAGE &     0.4405$\pm$0.1066      &    0.5439$\pm$ 0.0025    &   0.2589$\pm$0.1864     &    0.6416$\pm$0.0079      &   0.7589$\pm$0.0046     &   0.5949$\pm$0.0349     \\
& DR-GCN &     0.5523$\pm$0.0231      &    0.5921$\pm$0.0195     & 0.4038$\pm$0.0742   & 0.6488$\pm$0.0364  & 0.8295$\pm$0.0079  & 0.5357$\pm$0.1077  \\ \cline{2-8} 
& CARE-GNN  & 0.6075$\pm$0.0128    & 0.7713$\pm$0.0015  & 0.7023$\pm$0.0044 & 0.8875$\pm$0.0040   & 0.9398$\pm$0.0032 & 0.8848$\pm$0.0012 \\
& FRAUDRE   & 0.5841$\pm$0.0365 & 0.7427$\pm$0.0084 & 0.6654$\pm$0.0210 & 0.8806$\pm$0.0320  & 0.9272$\pm$0.0021 & 0.8808$\pm$0.0049 \\
& PC-GNN  & 0.6130$\pm$0.0083    & 0.7715$\pm$0.0005  & 0.7068$\pm$0.0015 & 0.8557$\pm$ 0.0227   & 0.9482$\pm$0.0034 & 0.8952$\pm$0.0044 \\
\hline
\multirow{2}{*}{Ablation}      
& DIGNN$_{\setminus S}$   & 0.5120$\pm$0.0027   & 0.6120$\pm$0.0067  & 0.5895$\pm$0.0010 &  0.7308$\pm$0.0064        &   0.8913$\pm$0.0020     &    0.8088$\pm$0.0042\\

& DIGNN$_{\setminus M}$   & 0.6994$\pm$0.0149    & 0.8389$\pm$0.0128  & 0.7348$\pm$0.0173 &  0.9186$\pm$0.0029        &   0.9645$\pm$0.0019     &    0.9195$\pm$0.0013    \\ \hline
         Ours             & DIGNN   & \textbf{0.7092}$\pm$\textbf{0.0025}    & \textbf{0.8526}$\pm$\textbf{0.0067}  & \textbf{0.7596}$\pm$\textbf{0.0105} &  \textbf{0.9189}$\pm$\textbf{0.0045}        &   \textbf{0.9729}$\pm$\textbf{0.0039}     &    \textbf{0.9281}$\pm$\textbf{0.0038}   \\ \hline
\end{tabular}}
\label{tab:performance}
\end{table*}

\begin{figure}[t]
\centering
\includegraphics[width=\linewidth]{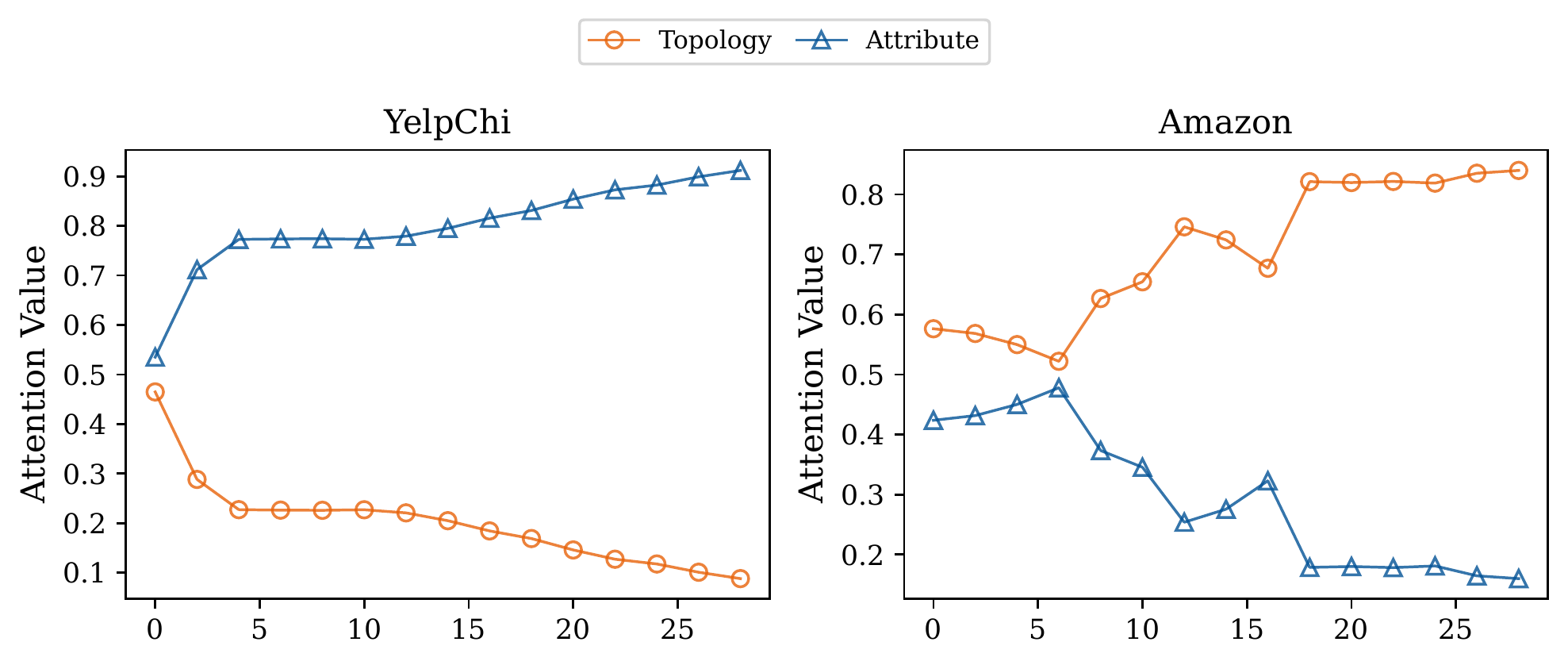}
\caption{The attention changing trends w.r.t epochs.}
\label{fig:attention}
\end{figure}

\subsubsection{Settings.} The parameters of DIGNN are optimized with Adam \cite{diederik2014adam} optimizer, the learning rate is set to 0.001, and weight decay is 0.0005, the training epochs are set to 50, the hidden dimension of node feature is set to 32, the scales of reconstruction learning rate of both topology and attribute are 0.05, the layer number of encoders are set to 2. The train, valid, and test ratio are set to be 40\%, 20\%, and 40\% respectively. We use Scikit-learn \cite{pedregosa2011scikit} to implement train-test split, and the imbalance ratio is consistent in three sets. It is worth noting that to alleviate the influence of class imbalance, we employ down-sampling to train DIGNN.

For GCN, GAT, and GraphSage, they suffer from the class imbalance and inconsistency problem, and will always predict normal (negative) samples. Therefor, we follow PC-GNN to utilize threshold-moving strategy, and the classification threshold is set to be 0.2 for YelpChi and Amazon. For CARE-GNN, FRAUDRE, PC-GNN, we use the parameters introduced by authors.

\subsubsection{Implementation.} Our model DIGNN is implemented in Pytorch 1.7.0 \cite{paszke2019pytorch}. For DIGNN, GCN, GAT, and GraphSage, we all implement them based on Pytorch Geometric 2.0.3 \cite{fey2019fast}. For CARE-GNN, FRAUDRE and PC-GNN, we carry out the source code provide by authors. All models are running on Python 3.8.12, 1 NVIDIA GeForce RTX 2080 GPU and 3.20 GHz Intel Xeon E5-2660 CPU.

\subsubsection{Metrics.} The fraud detection datasets display a skewed class distribution, so accuracy is not suitable to evaluate the effectiveness of fraud detection models. The evaluation metrics should have no bias to any class. Therefore, we use three common metrics, namely \textbf{F1-macro}, \textbf{AUC} and \textbf{GMean}. F1-macro is the unweighted mean of the F1-score of each class. AUC is the area under the ROC Curve. 
\begin{equation}
    \text{AUC}=\frac{\sum_{u\in\mathcal{U}^+}rank_u-\frac{|\mathcal{U}^+|\times(|\mathcal{U}^+|+1)}{2}}{|\mathcal{U}^+|\times|\mathcal{U}^-|}
    \nonumber
\end{equation}
Here, $\mathcal{U}^+$ and $\mathcal{U}^-$ denotes the minority and majority class set in the testing set, respectively. And $rank_u$ indicates the rank of node $u$ via the score of prediction. And GMean calculated the geometric mean of True Positive Rate (TPR) and True Negative Rate (TNR), it can be defined as,
\begin{equation}
    \text{GMean}=\sqrt{\text{TPR}\cdot\text{TNR}}=\sqrt{\frac{\text{TP}}{\text{TP}+\text{FN}}\cdot\frac{\text{TN}}{\text{TN}+\text{FP}}}.
    \nonumber
\end{equation}
The higher scores of this three metrics indicate the higher performance of the approaches.

\begin{figure*}[t]
\centering
\includegraphics[width=\textwidth]{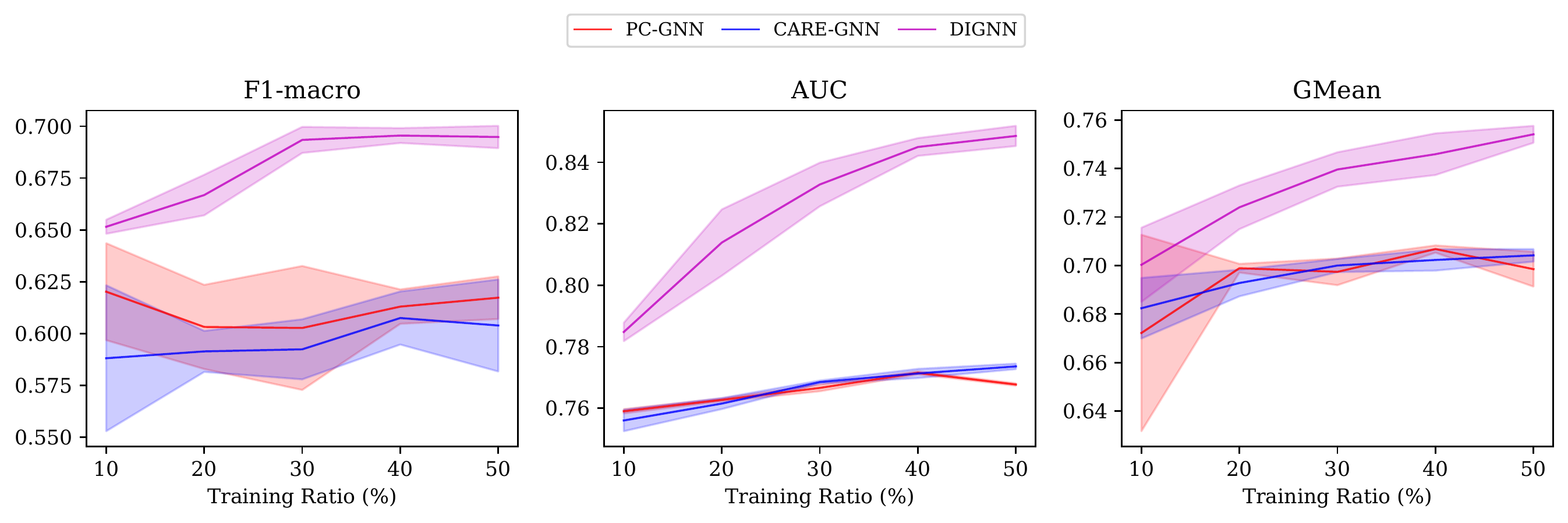}
\caption{Sensitivity analysis with respect to different training ratio on YelpChi dataset. The solid line represents the average score of 3 runs and the shadow indicates the standard deviation.}
\label{fig:training}
\end{figure*}

\begin{figure}[t]
\centering
\includegraphics[width=\linewidth]{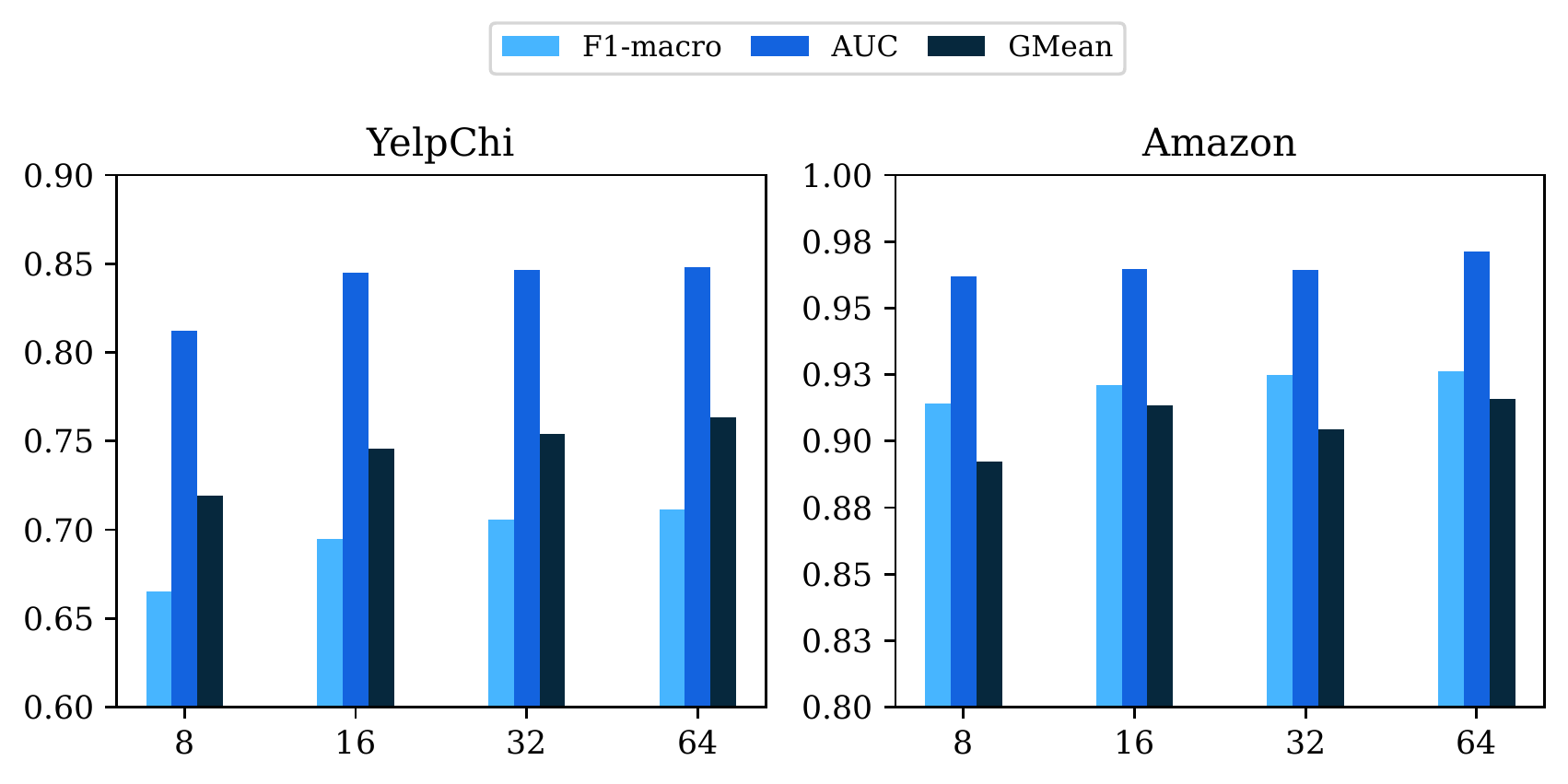}
\caption{Analysis of hidden dimension $d$.}
\label{fig:dimension}
\end{figure}

\begin{figure}[t]
\centering
\includegraphics[width=\linewidth]{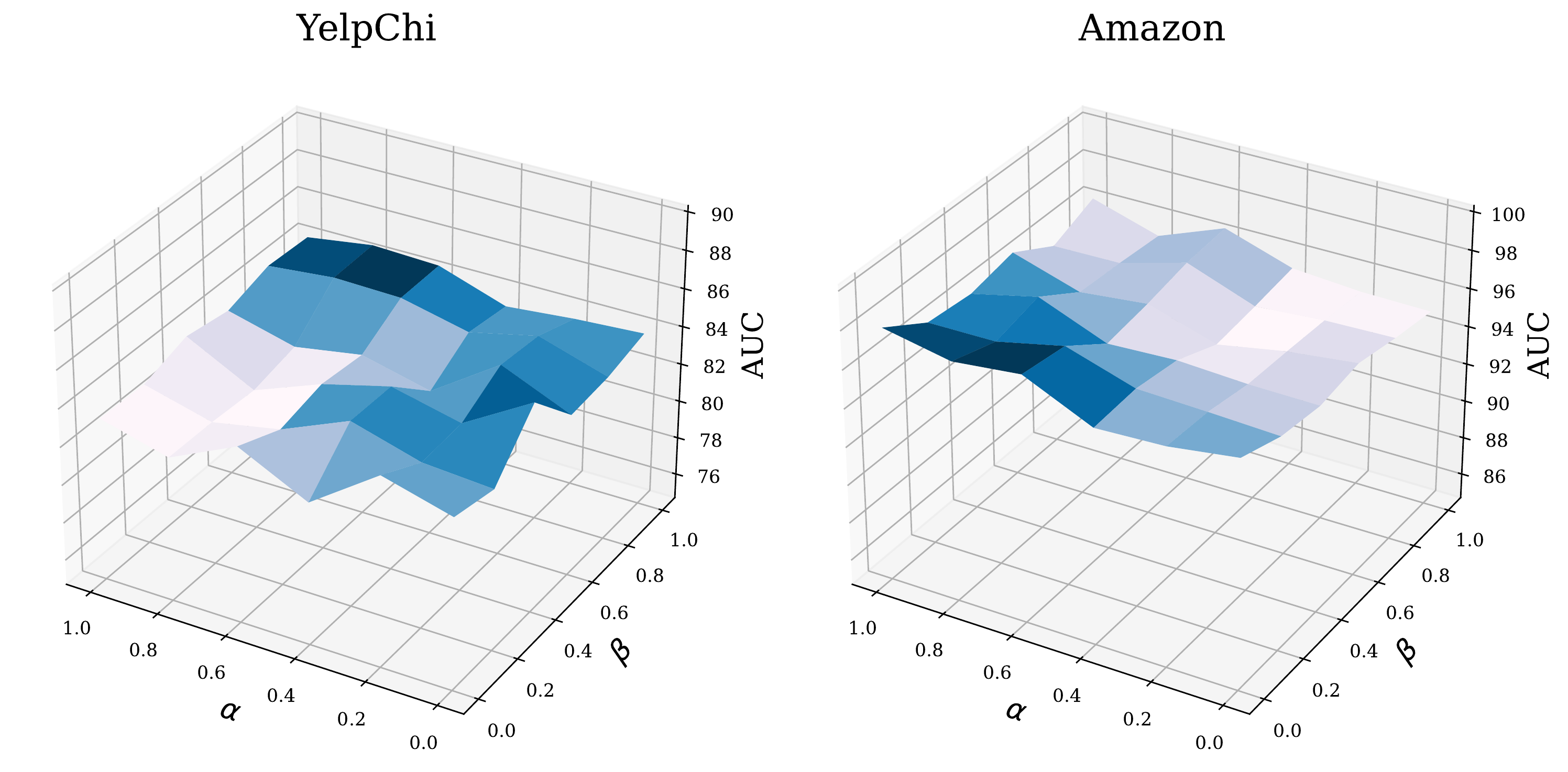}
\caption{AUC as the two hyper-parameters $\alpha$ and $\beta$ varying from 0 to 1.}
\label{fig:parameter}
\end{figure}

\begin{figure*}[t]
\centering
\includegraphics[width=\textwidth]{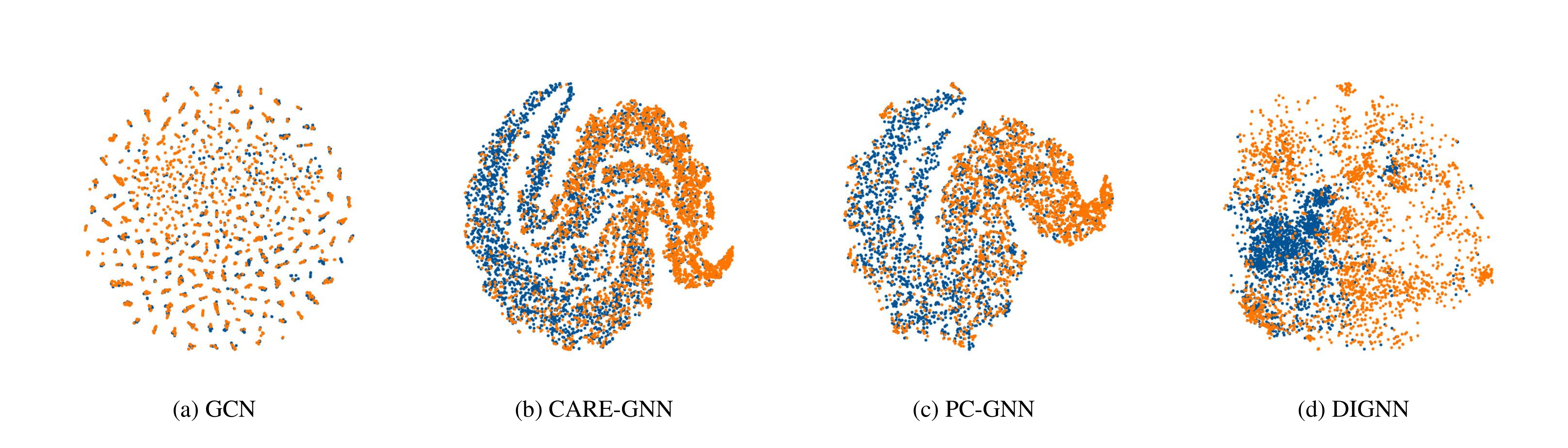}
\caption{Visualization of the learned node embeddings on YelpChi dataset.}
\label{fig:visilize}
\end{figure*}

\subsection{Analysis of Attention Mechanism (RQ1)}

To answer the RQ1, we analyze the attention values and visualize them for investigating whether the attention values learned by our model is meaningful. The attention changing trends are shown in \figref{fig:attention}. The x-axis is the number of training epochs and y-axis is the average attention value. With the training epoch increasing, the difference between the corresponding attention values of topology and attribute begin to be striking. We can observe that DIGNN pays more attention on attribute and topology on YelpChi and Amazon datasets respectively. It demonstrates our model has a strong capability to extract the task-relevant information from these two views.

\subsection{Performance Comparison (RQ2)}

To answer the RQ2, we compare the performance of DIGNN with state-of-the-art methods. The corresponding F1-macro, AUC and GMean scores are shown in \tabref{tab:performance}, we have the following two observations.

First, DIGNN significantly boosts the performance for all metrics on YelpChi and Amazon datasets than other SOTA baseline methods. In YelpChi dataset, our model obtains 9.62\%, 8.11\%, and 5.28\% improvement respectively in F1-macro, AUC and GMean. We can observe that PC-GNN outperforms other baselines in most metrics, but our model can still surpass it by a significant margin. In Amazon dataset, graph-based fraud detection methods have already achieved high performance and the increasing room is limited. But our model can still get improvements, with 3.14\% improvement in F1-macro, 2.47\% improvement in AUC and 3.29\% improvement in GMean. In addition, the relatively low standard deviation of DIGNN shows that our model is stable.

Second, the compared baseline methods can be divided into two groups, traditional MP-GNNs and graph-based fraud detection methods. GCN, GAT, GraphSAGE are tradition GNN models, and DR-GCN is designed for imbalanced node classes. They do not consider the inconsistency problem so that we can observe these models get poor performance on YelpChi and Amazon datasets. Because Amazon dataset has a more skewed label distribution (fraudsters only occupy 6.87\% of all samples), more intra-class edges which are beneficial to traditional MP-GNNs are appeared in graph. Thus these methods have a relatively satisfactory performance on Amazon. CARE-GNN and PC-GNN are graph-based fraud detection methods, they both sample neighbors according to similarity measure, which can alleviate inconsistency problem to a certain degree. Therefore, they can perform better on these two datasets. However, sampling neighbor strategy may discard a lot of information, and it can lead to sub-optimal results. Instead, our DIGNN model abandons this practice, and disentangles original graph into topological structure and node attributes, then processes them in parallel.

In general, DIGNN outperforms all baselines in F1-macro, AUC and GMean on YelpChi and Amazon datasets, which can demonstrate the effectiveness of our model.

\subsection{Ablation Study (RQ3)}

To answer the RQ3, we compare DIGNN with two corresponding variants DIGNN$_{\setminus S}$ and DIGNN$_{\setminus M}$. The results of two datasets are shown in \tabref{tab:performance}. We can observe that DIGNN surpasses its variants in most of metrics. For DIGNN$_{\setminus M}$, its overall performance on Yelpchi and Amazon is inferior to complete model, which verifies the effectiveness of our proposed mutual information objective. It is noting that the DIGNN$_{\setminus M}$ is on-par with DIGNN on Amazon dataset evaluated by F1-macro. This could attribute to smaller fraud rate of Amazon, thus DIGNN$_{\setminus M}$ pay more attention to majority class without the guidance of mutual information compared with DIGNN. For DIGNN$_{\setminus S}$, we can observe that DIGNN is evidently better than model without sampling strategy. We suppose it is caused by the noise information of the structure view. Sampling strategy  play a denoising effect on structural information to some extent.

\subsection{Sensitive Analysis (RQ4)}

To answer RQ4, we further evaluate the performance of DIGNN with respect to the training ratio, hidden dimension $d$ and hyperparameters $\alpha, \beta$. For training ratio, we vary the percentage of training nodes from 10\% to 50\%, and compare DIGNN with other two baselines, CARE-GNN and PC-GNN. \figref{fig:training} shows the performance of F1-macro, AUC and GMean on YelpChi dataset. We can observe that DIGNN always achieves best performance among the three models. When the training ratio is 10\%, DIGNN still performs better than PC-GNN training on 50\% samples. And DIGNN surpasses CARE-GNN and PC-GNN by a large margin in AUC.

For hidden dimension $d$, we study the performance of DIGNN with various hidden dimension number $d$ from 8 to 64. And the results are presented in \figref{fig:dimension}. With the increase of hidden dimension $d$, the performances improve first, but then start to grow slowly. It is relatively stable with respect to hidden dimension $d$.

For hyperparameters $\alpha$ and $\beta$, we vary these two hyperparameters from 0 to 1, and the corresponding results are shown in \figref{fig:parameter}. Considering the limit space, we only present AUC performance on YelpChi and Amazon datasets. It can be observed that the optimal selection of these two hyper-parameters varies greatly on the different datasets. In the YelpChi dataset, higher AUC performance can be achieved by selecting larger $\beta$ ($\beta\geq0.8$). And in the Amazon dataset, larger $\alpha$ ($\alpha\geq0.6$) and smaller $\beta$ ($\beta\leq0.4$) can get a better result.

\subsection{Visualization}

In order to show the effectiveness of different models more intuitively, we visualize the learned node embeddings on YelpChi dataset. Specifically, we compare our DIGNN model with the other three models. We select one traditional MP-GNN, GCN, and two graph-based fraud detection models, CARE-GNN and PC-GNN. Firstly, we use the 32-dimensional output embedding on the last layer of these models before softmax function. And then we employ the t-SNE \cite{van2008visualizing} to map the 32-dimensional embedding into 2-dimensional space for visualization. Because of the imbalanced class distribution, we randomly sample the same number of benign samples as fraudster samples for better visibility. The results of YelpChi are showed in \figref{fig:visilize}, and orange dots represent fraudsters, blue dots represent benign entities.

We can observe that due to the strong inductive bias of homophilic, GCN is disable to learn discriminative node embeddings. For CARE-GNN and PC-GNN, although they alleviate the inconsistency problem, they still fail to seperate the embeddings of fraudsters from that of the benign entities. Conversely, DIGNN achieves inter-class separation obviously, the overlap of the two kinds of nodes is relatively small. Consequently, it can verify the effectiveness of our proposed DIGNN model.


\section{Conclusion}

In this paper, we suggest that disentangling operation is beneficial to alleviate the inconsistency problem in fraud network. In order to decrease the conflict between topological structure and node attribute, we propose a simple yet effective model named DIGNN. It firstly disentangles the attribute fraud network into topology and attribute two views. Then DIGNN fuses two kinds of view information adaptively by attention mechanism, which can effectively extract task-relevant information. Moreover, we design a novel optimization objective to further reduce the entanglement between these two view-specific embeddings and maintain their semantic information. Experiment results demonstrate that DIGNN outperforms state-of-the-art methods on two real-world graph fraud detection datasets.

\section{acknowledge}

This work is jointly sponsored by National Natural Science Foundation of China (U19B2038, 62141608, 62206291) and CCF-AFSG Research Fund (20210001).

\bibliographystyle{IEEEtran}
\bibliography{reference}

\begin{thebibliography}{10}
\providecommand{\url}[1]{#1}
\csname url@samestyle\endcsname
\providecommand{\newblock}{\relax}
\providecommand{\bibinfo}[2]{#2}
\providecommand{\BIBentrySTDinterwordspacing}{\spaceskip=0pt\relax}
\providecommand{\BIBentryALTinterwordstretchfactor}{4}
\providecommand{\BIBentryALTinterwordspacing}{\spaceskip=\fontdimen2\font plus
\BIBentryALTinterwordstretchfactor\fontdimen3\font minus
  \fontdimen4\font\relax}
\providecommand{\BIBforeignlanguage}[2]{{%
\expandafter\ifx\csname l@#1\endcsname\relax
\typeout{** WARNING: IEEEtran.bst: No hyphenation pattern has been}%
\typeout{** loaded for the language `#1'. Using the pattern for}%
\typeout{** the default language instead.}%
\else
\language=\csname l@#1\endcsname
\fi
#2}}
\providecommand{\BIBdecl}{\relax}
\BIBdecl

\bibitem{li2019spam}
A.~Li, Z.~Qin, R.~Liu, Y.~Yang, and D.~Li, ``Spam review detection with graph
  convolutional networks,'' in \emph{Proceedings of the 28th ACM International
  Conference on Information and Knowledge Management}, 2019, pp. 2703--2711.

\bibitem{dou2021user}
Y.~Dou, K.~Shu, C.~Xia, P.~S. Yu, and L.~Sun, ``User preference-aware fake news
  detection,'' in \emph{Proceedings of the 44th International ACM SIGIR
  Conference on Research and Development in Information Retrieval}, 2021, pp.
  2051--2055.

\bibitem{xu2022mining}
W.~Xu, J.~Wu, Q.~Liu, S.~Wu, and L.~Wang, ``Mining fine-grained semantics via
  graph neural networks for evidence-based fake news detection,'' \emph{arXiv
  preprint arXiv:2201.06885}, 2022.

\bibitem{deng2022markov}
L.~Deng, C.~Wu, D.~Lian, Y.~Wu, and E.~Chen, ``Markov-driven graph
  convolutional networksfor social spammer detection,'' \emph{IEEE Transactions
  on Knowledge and Data Engineering}, 2022.

\bibitem{wang2019semi}
D.~Wang, J.~Lin, P.~Cui, Q.~Jia, Z.~Wang, Y.~Fang, Q.~Yu, J.~Zhou, S.~Yang, and
  Y.~Qi, ``A semi-supervised graph attentive network for financial fraud
  detection,'' in \emph{2019 IEEE International Conference on Data Mining
  (ICDM)}.\hskip 1em plus 0.5em minus 0.4em\relax IEEE, 2019, pp. 598--607.

\bibitem{liu2021pick}
Y.~Liu, X.~Ao, Z.~Qin, J.~Chi, J.~Feng, H.~Yang, and Q.~He, ``Pick and choose:
  a gnn-based imbalanced learning approach for fraud detection,'' in
  \emph{Proceedings of the Web Conference 2021}, 2021, pp. 3168--3177.

\bibitem{kipf2016semi}
T.~N. Kipf and M.~Welling, ``Semi-supervised classification with graph
  convolutional networks,'' \emph{arXiv preprint arXiv:1609.02907}, 2016.

\bibitem{hamilton2017inductive}
W.~Hamilton, Z.~Ying, and J.~Leskovec, ``Inductive representation learning on
  large graphs,'' \emph{Advances in neural information processing systems},
  vol.~30, 2017.

\bibitem{velickovic2017graph}
P.~Velickovic, G.~Cucurull, A.~Casanova, A.~Romero, P.~Lio, and Y.~Bengio,
  ``Graph attention networks,'' \emph{stat}, vol. 1050, p.~20, 2017.

\bibitem{nt2019revisiting}
H.~Nt and T.~Maehara, ``Revisiting graph neural networks: All we have is
  low-pass filters,'' \emph{arXiv preprint arXiv:1905.09550}, 2019.

\bibitem{liu2020alleviating}
Z.~Liu, Y.~Dou, P.~S. Yu, Y.~Deng, and H.~Peng, ``Alleviating the inconsistency
  problem of applying graph neural network to fraud detection,'' in
  \emph{Proceedings of the 43rd international ACM SIGIR conference on research
  and development in information retrieval}, 2020, pp. 1569--1572.

\bibitem{ma2021unified}
Y.~Ma, X.~Liu, T.~Zhao, Y.~Liu, J.~Tang, and N.~Shah, ``A unified view on graph
  neural networks as graph signal denoising,'' in \emph{Proceedings of the 30th
  ACM International Conference on Information \& Knowledge Management}, 2021,
  pp. 1202--1211.

\bibitem{yang2022graph}
L.~Yang, W.~Zhou, W.~Peng, B.~Niu, J.~Gu, C.~Wang, X.~Cao, and D.~He, ``Graph
  neural networks beyond compromise between attribute and topology,'' 2022.

\bibitem{wang2020gcn}
X.~Wang, M.~Zhu, D.~Bo, P.~Cui, C.~Shi, and J.~Pei, ``Am-gcn: Adaptive
  multi-channel graph convolutional networks,'' in \emph{Proceedings of the
  26th ACM SIGKDD International conference on knowledge discovery \& data
  mining}, 2020, pp. 1243--1253.

\bibitem{dou2020enhancing}
Y.~Dou, Z.~Liu, L.~Sun, Y.~Deng, H.~Peng, and P.~S. Yu, ``Enhancing graph
  neural network-based fraud detectors against camouflaged fraudsters,'' in
  \emph{Proceedings of the 29th ACM International Conference on Information \&
  Knowledge Management}, 2020, pp. 315--324.

\bibitem{wang2021decoupling}
Y.~Wang, J.~Zhang, S.~Guo, H.~Yin, C.~Li, and H.~Chen, ``Decoupling
  representation learning and classification for gnn-based anomaly detection,''
  in \emph{Proceedings of the 44th International ACM SIGIR Conference on
  Research and Development in Information Retrieval}, 2021, pp. 1239--1248.

\bibitem{ao2021temporal}
X.~Ao, Y.~Liu, Z.~Qin, Y.~Sun, and Q.~He, ``Temporal high-order proximity aware
  behavior analysis on ethereum,'' \emph{World Wide Web}, vol.~24, no.~5, pp.
  1565--1585, 2021.

\bibitem{liang2021credit}
T.~Liang, G.~Zeng, Q.~Zhong, J.~Chi, J.~Feng, X.~Ao, and J.~Tang, ``Credit risk
  and limits forecasting in e-commerce consumer lending service via
  multi-view-aware mixture-of-experts nets,'' in \emph{Proceedings of the 14th
  ACM international conference on web search and data mining}, 2021, pp.
  229--237.

\bibitem{zhang2022efraudcom}
G.~Zhang, Z.~Li, J.~Huang, J.~Wu, C.~Zhou, J.~Yang, and J.~Gao, ``efraudcom: An
  e-commerce fraud detection system via competitive graph neural networks,''
  \emph{ACM Transactions on Information Systems (TOIS)}, vol.~40, no.~3, pp.
  1--29, 2022.

\bibitem{ma2021comprehensive}
X.~Ma, J.~Wu, S.~Xue, J.~Yang, C.~Zhou, Q.~Z. Sheng, H.~Xiong, and L.~Akoglu,
  ``A comprehensive survey on graph anomaly detection with deep learning,''
  \emph{IEEE Transactions on Knowledge and Data Engineering}, 2021.

\bibitem{zhang2021fraudre}
G.~Zhang, J.~Wu, J.~Yang, A.~Beheshti, S.~Xue, C.~Zhou, and Q.~Z. Sheng,
  ``Fraudre: Fraud detection dual-resistant to graph inconsistency and
  imbalance,'' in \emph{2021 IEEE International Conference on Data Mining
  (ICDM)}.\hskip 1em plus 0.5em minus 0.4em\relax IEEE, 2021, pp. 867--876.

\bibitem{liu2021intention}
C.~Liu, L.~Sun, X.~Ao, J.~Feng, Q.~He, and H.~Yang, ``Intention-aware
  heterogeneous graph attention networks for fraud transactions detection,'' in
  \emph{Proceedings of the 27th ACM SIGKDD Conference on Knowledge Discovery \&
  Data Mining}, 2021, pp. 3280--3288.

\bibitem{liu2021self}
C.~Liu, L.~Wen, Z.~Kang, G.~Luo, and L.~Tian, ``Self-supervised consensus
  representation learning for attributed graph,'' in \emph{Proceedings of the
  29th ACM International Conference on Multimedia}, 2021, pp. 2654--2662.

\bibitem{lim2021large}
D.~Lim, F.~Hohne, X.~Li, S.~L. Huang, V.~Gupta, O.~Bhalerao, and S.~N. Lim,
  ``Large scale learning on non-homophilous graphs: New benchmarks and strong
  simple methods,'' \emph{Advances in Neural Information Processing Systems},
  vol.~34, 2021.

\bibitem{Federici2020}
M.~Federici, A.~Dutta, P.~Forr{\'e}, N.~Kushman, and Z.~Akata, ``Learning
  robust representations via multi-view information bottleneck,'' \emph{arXiv
  preprint arXiv:2002.07017}, 2020.

\bibitem{bao2021}
F.~Bao, ``Disentangled variational information bottleneck for multiview
  representation learning,'' in \emph{CAAI International Conference on
  Artificial Intelligence}.\hskip 1em plus 0.5em minus 0.4em\relax Springer,
  2021, pp. 91--102.

\bibitem{wan2021}
Z.~Wan, C.~Zhang, P.~Zhu, and Q.~Hu, ``Multi-view information-bottleneck
  representation learning,'' in \emph{Proceedings of the AAAI Conference on
  Artificial Intelligence}, vol.~35, no.~11, 2021, pp. 10\,085--10\,092.

\bibitem{Belghazi2018}
M.~I. Belghazi, A.~Baratin, S.~Rajeshwar, S.~Ozair, Y.~Bengio, A.~Courville,
  and D.~Hjelm, ``Mutual information neural estimation,'' in
  \emph{International conference on machine learning}.\hskip 1em plus 0.5em
  minus 0.4em\relax PMLR, 2018.

\bibitem{Cover1991}
T.~M. Cover, J.~A. Thomas \emph{et~al.}, ``Entropy, relative entropy and mutual
  information,'' \emph{Elements of information theory}, vol.~2, no.~1, pp.
  12--13, 1991.

\bibitem{vincent2010stacked}
P.~Vincent, H.~Larochelle, I.~Lajoie, Y.~Bengio, P.-A. Manzagol, and L.~Bottou,
  ``Stacked denoising autoencoders: Learning useful representations in a deep
  network with a local denoising criterion.'' \emph{Journal of machine learning
  research}, vol.~11, no.~12, 2010.

\bibitem{kingma2013auto}
D.~P. Kingma and M.~Welling, ``Auto-encoding variational bayes,'' \emph{arXiv
  preprint arXiv:1312.6114}, 2013.

\bibitem{rayana2015collective}
S.~Rayana and L.~Akoglu, ``Collective opinion spam detection: Bridging review
  networks and metadata,'' in \emph{Proceedings of the 21th acm sigkdd
  international conference on knowledge discovery and data mining}, 2015, pp.
  985--994.

\bibitem{mcauley2013amateurs}
J.~J. McAuley and J.~Leskovec, ``From amateurs to connoisseurs: modeling the
  evolution of user expertise through online reviews,'' in \emph{Proceedings of
  the 22nd international conference on World Wide Web}, 2013, pp. 897--908.

\bibitem{shi2020multi}
M.~Shi, Y.~Tang, X.~Zhu, D.~Wilson, and J.~Liu, ``Multi-class imbalanced graph
  convolutional network learning,'' in \emph{Proceedings of the Twenty-Ninth
  International Joint Conference on Artificial Intelligence (IJCAI-20)}, 2020.

\bibitem{diederik2014adam}
K.~Diederik, B.~Jimmy \emph{et~al.}, ``Adam: A method for stochastic
  optimization,'' \emph{arXiv preprint arXiv:1412.6980}, pp. 273--297, 2014.

\bibitem{pedregosa2011scikit}
F.~Pedregosa, G.~Varoquaux, A.~Gramfort, V.~Michel, B.~Thirion, O.~Grisel,
  M.~Blondel, P.~Prettenhofer, R.~Weiss, V.~Dubourg \emph{et~al.},
  ``Scikit-learn: Machine learning in python,'' \emph{the Journal of machine
  Learning research}, vol.~12, pp. 2825--2830, 2011.

\bibitem{paszke2019pytorch}
A.~Paszke, S.~Gross, F.~Massa, A.~Lerer, J.~Bradbury, G.~Chanan, T.~Killeen,
  Z.~Lin, N.~Gimelshein, L.~Antiga \emph{et~al.}, ``Pytorch: An imperative
  style, high-performance deep learning library,'' \emph{Advances in neural
  information processing systems}, vol.~32, 2019.

\bibitem{fey2019fast}
M.~Fey and J.~E. Lenssen, ``Fast graph representation learning with pytorch
  geometric,'' \emph{arXiv preprint arXiv:1903.02428}, 2019.

\bibitem{van2008visualizing}
L.~Van~der Maaten and G.~Hinton, ``Visualizing data using t-sne.''
  \emph{Journal of machine learning research}, vol.~9, no.~11, 2008.

\end{thebibliography}

\end{document}